%% file: root.tex
\documentclass[letterpaper, 10 pt, conference]{ieeeconf}  
\usepackage{xcolor}
\usepackage{hyperref}
\usepackage{amsfonts}
\usepackage{amsmath}
\usepackage{cite}
\usepackage{adjustbox}
\usepackage{hyperref}

\usepackage{etoolbox,siunitx,booktabs,threeparttable,multirow,array,graphicx}
\usepackage[normalem]{ulem}

\robustify\bfseries
\robustify\uline
\robustify\tnote

\newcommand{\model}{GrAMMI} 
\newcommand{\draft}[1]{\textcolor{black}{#1}}

\IEEEoverridecommandlockouts                              

\usepackage{amssymb}  

\title{\LARGE \bf
Learning Models of Adversarial Agent Behavior under Partial Observability
}

 \author{Sean Ye$^{1,\dagger}$, Manisha Natarajan$^{1,\dagger}$, Zixuan Wu$^{1,\dagger}$, Rohan Paleja$^{1}$, Letian Chen$^{1}$,  and Matthew C. Gombolay$^{1}$
 \thanks{*This work supported in part by Office of Naval Research (ONR) under grant numbers N00014-19-1-2076, N00014-22-1-2834, and N00173-21-1-G009, the National Science Foundation under grant CNS-2219755, and MIT Lincoln Laboratory grant number 7000437192.}
 \thanks{$\dagger$ Equal contribution}
 \thanks{$^{1}$All authors are associated with the Institute of Robotics and Intelligent Machines (IRIM), Georgia Institute of Technology, Atlanta, GA, USA.}%
 \thanks{Corresponding Author: Sean Ye,
 		{\tt\small seancye@gatech.edu}}
 }

\begin{document}

\maketitle
\thispagestyle{empty}
\pagestyle{empty}



\begin{abstract}

The need for opponent modeling and tracking arises in several real-world scenarios, such as professional sports, video game design, and drug-trafficking interdiction. In this work, we present \underline{Gr}aph based \underline{A}dversarial \underline{M}odeling with \underline{M}utal \underline{I}nformation (\model) for modeling the behavior of an adversarial opponent agent. \model{} is a novel graph neural network (GNN) based approach that uses mutual information maximization as an auxiliary objective to predict the current and future states of an adversarial opponent with partial observability. To evaluate \model{}, we design two large-scale, pursuit-evasion domains inspired by real-world scenarios, where a team of heterogeneous agents is tasked with tracking and interdicting a single adversarial agent, and the adversarial agent must evade detection while achieving its own objectives. With the mutual information formulation, \model{} outperforms all baselines in both domains and achieves $31.68$\% higher log-likelihood on average for future adversarial state predictions across both domains.

\end{abstract}

\section{INTRODUCTION}


According to the World Drug Report from the United Nations Office on Drugs and Crime (UNODC), over 39 million individuals have been impacted by illicit drugs, leading to various disorders, including HIV infection, hepatitis-related liver diseases, overdose, and premature death in 2022 \cite{UNODC}. To safeguard the health and well-being of people across the globe, it is imperative to develop advanced drug traffic interdiction strategies to aid law enforcement.

Drug traffic interdiction can be formulated as an opponent modeling problem. We define opponent modeling as \textit{the ability to use prior knowledge to predict an opponent's behavior, whose internal states are not fully observable}. The need for opponent modeling is not limited to drug traffic interdiction and can arise in several real-world scenarios, such as search-and-rescue, border patrol, professional sports, or military surveillance, where an intelligent, evasive target must be monitored under partial observability \cite{nashed2022survey} by a team of surveilling agents. In this work, we propose a novel deep learning framework \model{} (\textbf{Gr}aph based \textbf{A}dversarial \textbf{M}odeling with \textbf{M}utal \textbf{I}nformation) for opponent modeling in challenging, large-scale domains inspired by various real-world scenarios.

Opponent modeling and tracking involve two key synergistic components: 1) the use of prior observations to infer a model of the opponent's behavior and 2) leveraging this model to observe and track the opponent actively, aiming to gather more observations of the opponent's behavior. 
This is a highly challenging problem as we only have access to a limited number of observations of the adversary currently being tracked (dependent on our tracking ability), partial information is provided upon observation of an adversary (e.g., the adversary's latent intentions are hidden information), and an intelligent adversary will change its behavior upon detection to minimize future detection.
As adversaries may have multiple possible destinations, latent preferences across destinations, and adapt such preferences upon detection, uncertainties across a large state-space must be effectively maintained and updated during observation and lack-of observation. The ability to maintain a multi-hypothesis belief over the adversary is pivotal for effective opponent models.

\begin{figure}
    \centering
    \includegraphics[width=0.5\textwidth, keepaspectratio]{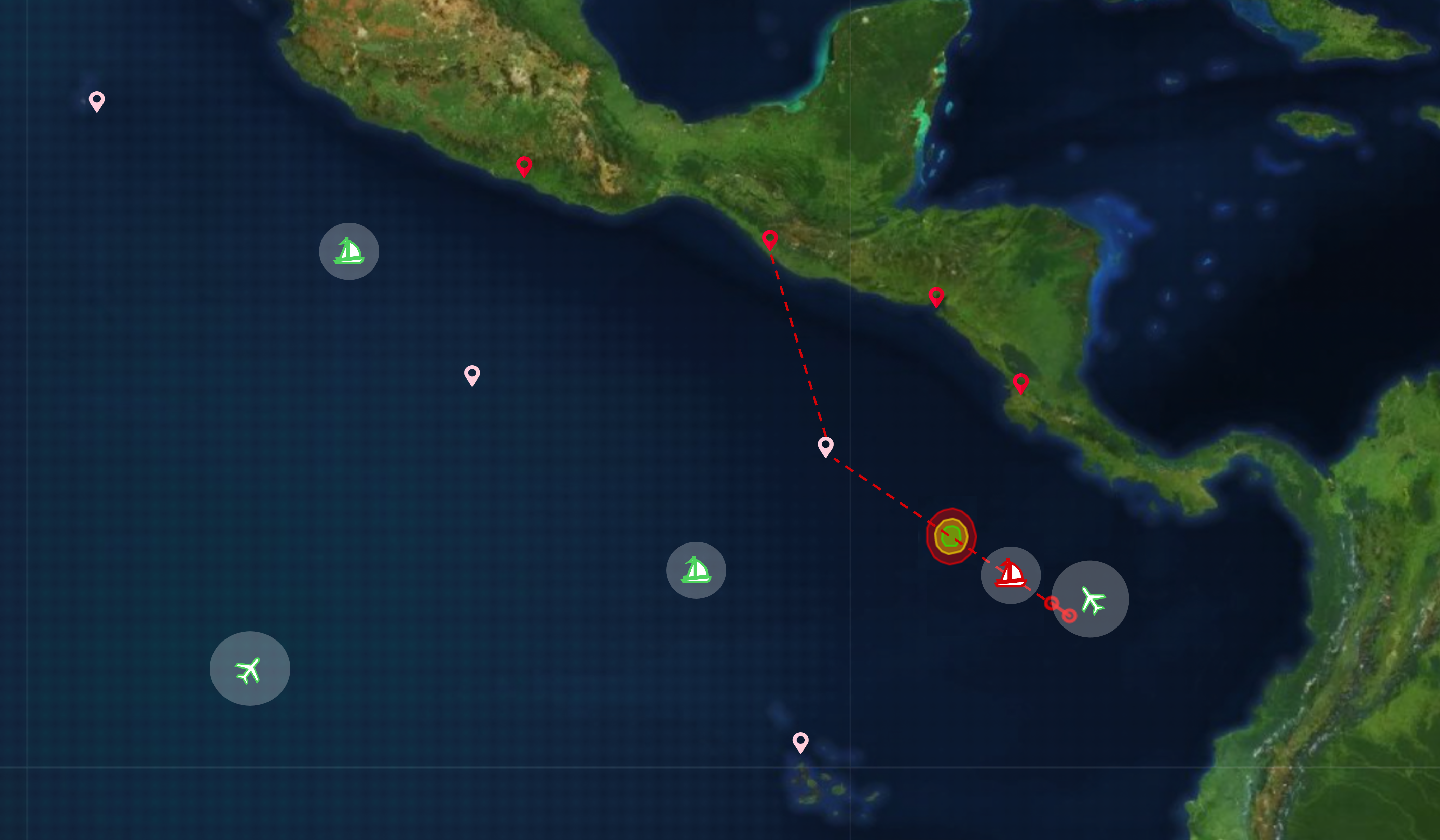}
    \caption{Narco-Traffic Interdiction: An adversarial opponent (red vessel) is tracked by a team of heterogeneous pursuit agents (shown in green) under partial observability.  The trajectory forecast T steps in the future is shown via multi-colored 2D Gaussians}
    \label{fig:narco}
\end{figure}

Traditionally, target tracking (adversarial and otherwise) has been dominated by classical filtering methods such as Kalman Filters \cite{chen2000mixture, leven2009unscented} and Particle Filters \cite{djuric2008target, rao2013visual, jia2016target}, where a dynamic motion model of the target is assumed to be known \cite{li2003}. Such model-based filtering approaches require the target's perspective (i.e., access to true target states) and tend to work well only if the dynamics model of the target is known or can be estimated accurately. In our work, we focus on domains that do not have access to the true states of the opponent nor the opponent's true motion model, and hence we choose not to rely on model-based approaches. Inspired by the recent success of model-free approaches for agent modeling problems in other domains such as visual tracking and trajectory forecasting \cite{lu2014,ivanovic2022propagating},
we develop a graph-based, model-free approach, \model{}, that uses an auxiliary \emph{mutual information} objective for tracking multiple hypotheses of the opponent's states under partial observability.


While most prior works in opponent modeling have been restricted to small grid worlds \cite{singh2018learning}, we seek to perform adversarial opponent tracking in large state spaces, \textcolor{black}{100$\times$} the size of those considered in prior work. Hence, we create two novel, open-source domains -- \textbf{\textit{Narco Traffic Interdiction}}  and \textbf{\textit{Prison Escape}}, inspired by real-world scenarios. Narco Traffic Interdiction is designed to address illegal drug trafficking by sea (mainly in shipping containers), which is estimated to be rapidly growing and  accounted for over 90 percent of cocaine seized globally in 2021 \cite{UNODC}. Prison Escape is designed as a complex pursuit-evasion domain inspired by military surveillance problems, where a highly intelligent adversary is capable of adapting its behaviors to evade the tracking agents. Each domain is designed as a grid-world environment with a grid size in the order of $\sim 10^3 \times 10^3$. The state space within each domain is large: $O((m \times n)^k)$, where $m \times n$ is the grid size, and $k$ is the total number of agents in the environment. In both domains, a single adversarial opponent is pursued by a team of heterogeneous tracker agents (i.e., agents with different capabilities, such as speed and detection accuracy), and these agents must utilize their heterogeneous capabilities to coordinate and best track the adversary. Furthermore, the detection ability of the tracker agents can degrade across the terrain, e.g., denser forest results in lower detection ability. Despite the challenges posed in these domains, we show that our proposed approach \model{} can outperform all baselines across a variety of evaluation metrics for both domains on varying levels of difficulty.

\vspace{5pt}




\noindent \textbf{Contributions:} Our key contributions are two-fold. 
\begin{itemize}
    \item First, we propose \model{}, a deep-learning architecture that predicts the present and future states of an adversarial agent using a mutual information formulation and multi-agent graph communication. Our results demonstrate that incorporating a mutual information maximization objective improves the estimation of an adversary's location by disentangling latent embeddings to account for multimodal hypotheses explicitly. \model{} achieves 40.54\% and 18.39\% higher log-likelihood on average across all baselines for the Prison Escape scenario and Narco Traffic Interdiction domains, respectively.
    \item Second, we introduce two new adversarial domains with continuous action spaces, where an intelligent adversary is pursued by a team of heterogeneous agents. We open-source these domains at \url{https://github.com/CORE-Robotics-Lab/Opponent-Modeling} to motivate further research in opponent modeling.
\end{itemize}
    
    

\section{Related Works}
\label{sec:related_work}



 \color{black}{Reasoning about the goals, beliefs, and behaviors of opponents can enable agents to develop effective strategies for succeeding in opponent modeling settings. In this work, we propose to use neural networks to learn the representations of opponents. Most prior works in opponent modeling assume constant access to opponent states or observations during both training and inference to learn predictive models of opponent behavior \cite{he2016opponent, raileanu2018modeling, grover2018learning}. Such an assumption is unrealistic due to the fact that the opponent is non-cooperative and that each agent is equipped with a limited field of view. Thus, the agents may not know the true location of the opponent at all times. Only recent work by Papoudakis et al. forgoes this assumption by eliminating access to opponent information during inference \cite{papoudakis2020variational}. However, they simply use the latent representation from the variational autoencoder (trained with full state information) for downstream RL tasks. Further, prior works are limited to predicting opponent behavior over a short horizon in small domains \cite{grover2018learning, papoudakis2020variational}. \emph{To the best of our knowledge, we are the first to look at opponent modeling under partial observability and limited observability (i.e., with no access to opponent information for model inputs) for complex, large-scale domains. Furthermore, we evaluate the performance of our proposed approach in predicting the future states of an adversarial opponent for both short and long horizons. }}

\vspace{4pt}

\color{black}

Adversarial Opponent Modeling can be framed as a Partially Observable Markov Game (POMG), where we utilize limited observations from a set of tracking agents to predict opponent behavior. This is highly similar to the problem of imitation learning, where observations or states can be used to infer a mapping from user states/observations to actions. 
Multimodality in opponent modeling may arise from an adversary's latent preferences, a lack of observations resulting in an expansion across possible locations over time, or due to the adversary employing multimodal evasive behaviors. Prior work in multimodal imitation learning address heterogeneity \cite{infoGAIL, paleja2020interpretable, hausman2017multi, tangkaratt2020variational} and suboptimality \cite{wu2019imitation, chen2021learning, schrum2022mind} of expert demonstrations (state-action pairs), by typically employing deep generative models such as Generative Adversarial Networks (GANs) or Variational Autoencoders (VAEs) to discover salient latent factors that can account for multimodality. Such approaches utilize the full state information of the agent that they are modeling. In contrast, \emph{we aim to learn multimodal predictions for adversary states under partial and limited observability, i.e., we only have access to sparse, intermittent observations of the opponent.}


\section{Background}
\label{sec:background}
\subsection{Partially Observable Markov Game}


We define the opponent modeling problem as a Partially Observable Markov Game (POMG), which consists of a set of states $\mathcal{S}$, a set of private agent observations ${\mathcal{O}_1, \mathcal{O}_2, \ldots, \mathcal{O}_N}$, a set of actions ${\mathcal{A}_1, \mathcal{A}_2, \ldots, \mathcal{A}_N}$, and a transition function $\mathcal{T} : \mathcal{S} \times \mathcal{A}_1 \times \ldots \times \mathcal{A}_N \mapsto \mathcal{S}$ for N-agents. At each time step $t$, agents receive an observation $O_i^t\in\mathcal{O}_i$, choose an action $a_i^t \in \mathcal{A}_i$, and receive a reward $r_i^t$ based on the reward function $R: \mathcal{S} \times \mathcal{A}_i \mapsto \mathbb{R}$. The initial state is drawn from an initial state distribution $\rho$. Opponent modeling involves two teams -- tracking agents ($\mathbf{A}^+$) and adversaries ($\mathbf{A}^-$). In both our domains, we have a single adversary being tracked by a team of heterogeneous agents. Thus, our approach, \model{}, learns a mapping from a history of observations of the tracker agents, $\mathcal{O}^{t-H:t}_{i \in A^+}$ to the future state of the adversary, $S^{t+T}_{i \in A^-}$, where $H$ refers to a length of history and $T$ refers to a timepoint in the future.


\subsection{Graph Neural Networks}
Graph Neural Networks (GNNs) allow deep learning approaches to learn from data with graph structures \cite{wu2020comprehensive}. Graphs are represented as $G = (V, E)$, where $V$ is the set of nodes and $E$ is the set of edges. For every edge $e_{ij} \in E$, $e_{ij} = (v_i, v_j)$ where $v_i$ is the start node and $v_j$ is the end node. Graphs may contain both node and edge features where each node $v_i \in V$ has a corresponding vector of $x_{v_i} \in \mathbb{R}^D$. 

GNN layers utilize message passing to aggregate feature vectors from neighboring nodes in the graph. The update rule for learning node representations in GNNs is described by Equation (\ref{eq:GNN})
\begin{equation}
\label{eq:GNN}
    h_i^{(l)} = \sigma \left(\sum\limits_{j \in N_i} \frac{1}{\sqrt{d_id_j}} (h_j^{(l-1)}W^{(l)}) \right)
\end{equation}
In Equation (\ref{eq:GNN}), $h_i^{(l)}$ represents the features of $v_i$ at layer $l$. $N_i$ is the set of neighboring nodes for $v_i$, $d_i = |N_i|$ is the degree of node $v_i$, $W^{(l)}$ is a learned weight parameter for layer $l$, and $\sigma(\cdot)$ is a non-linear activation function. We utilize GNNS to model team interaction across tracker agents in our domains.

\section{Method}
\label{sec:method}

\begin{figure*}[ht]
\centering
    \includegraphics[width=0.82\textwidth]{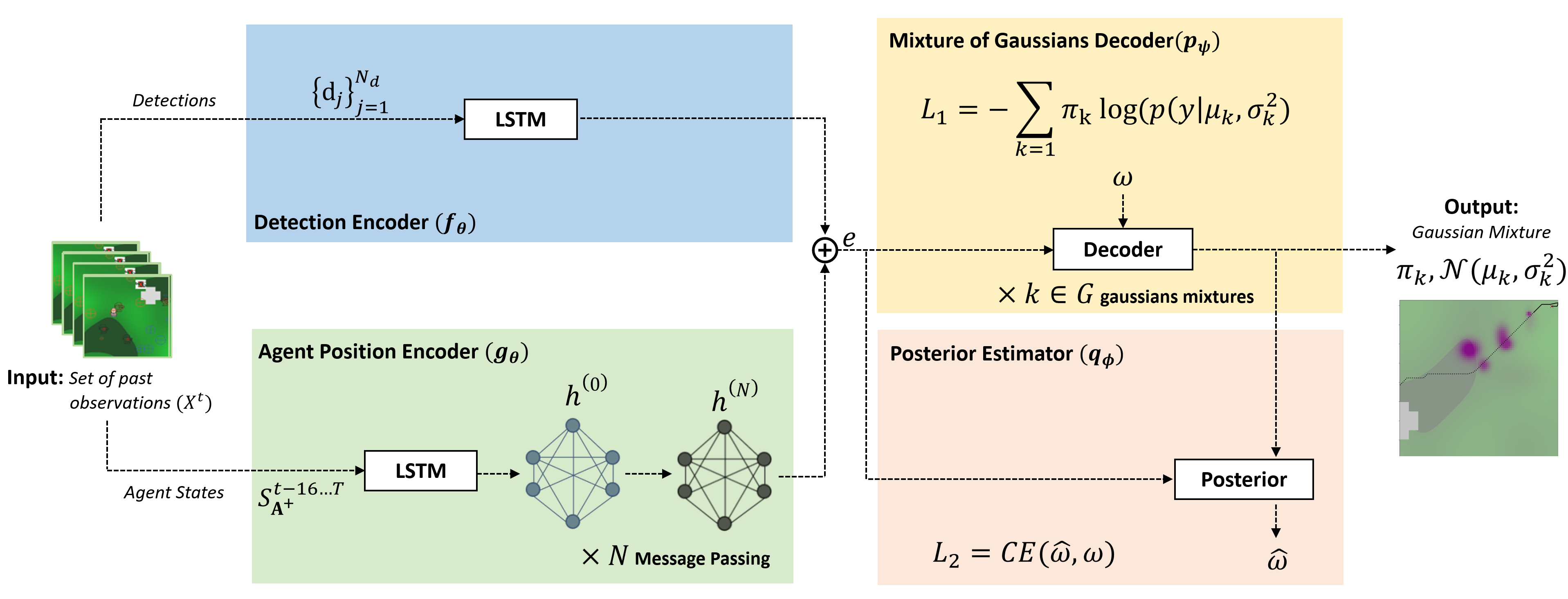}
    
    \caption{Our proposed architecture \model{} uses graph neural networks with mutual information maximization of Gaussian components to predict the location of an adversarial target for large-scale domains with partial observability. The data buffer consists of trajectories collected by online interaction within our two domains.}
    \label{fig:architecture}
\end{figure*}

\subsection{Problem Formulation}
In this work, we aim to generate the current and future state distributions for a dynamic adversarial opponent, from the observations of a team of $N$ heterogeneous tracker agents $\{A_i, i \in \mathbf{A^+}\}$, each of which has a semantic type $C_i$ (for instance, camera, search party, helicopter). We refer to the team of tracker agents as the \textit{blue} team ($\mathbf{A^+}$), and the \textit{red} team ($\mathbf{A^-}$) refers to the adversarial opponent being tracked. We may use the terms blue team for tracker agents and red team for the adversarial opponent interchangeably.

We model the predicted states of the adversarial agent as a multimodal probability distribution over the space of the map to account for the possibility of multiple distinct futures for the adversary, given that we only have access to limited, partial observations of the adversary from the blue agents. 

At time $t$, each blue agent $A_i$'s state $S_i^t \in \mathbb{R}^D$ consists of its current location ($x_i^t, y_i^t$), a one-hot encoding for the agent's semantic type $C_i$, and the current timestep $t$ in the episode. Each blue agent's observation $O_i^t$ $\in \mathbb{R}^3$ consists of a flag to denote whether or not the blue agent detected the adversary and the detected location of the adversary ($b_i^t, \hat{x}_i^t, \hat{y}_i^t$),  We assume a centralized communication framework, where all blue team agents can communicate their observations to each other at every timestep and, thus, maintain a common opponent detection history $\lbrace d_j \rbrace _{j=1}^{N_{d}}$, where $N_{d}$ is the number of detections from the start of the episode, and $d_j$ is the detected location of the opponent (i.e., ($\hat{x}^t, \hat{y}^t$)) at time $t=j$ by any agent in $\mathbf{A^+}$. Further, we assume that all detections of the opponent are accurate, i.e., there is no discordance in the opponent's estimated location when multiple blue team agents observe the opponent. 

Given a history of blue team states for the previous $H$ timesteps, $S_{\mathbf{A^+}}^{(t-H:t)} \in \mathbb{R}^{(H+1) \times N \times D}$,  and the shared detection history, $\{d_j\}_{j=1}^{N_d}$, our goal is to \draft{perform} filtering ($T=0$) and prediction ($T>0$) of the adversary $\mathcal{Y}^t = {S}_{\mathbf{A^-}}^{(t+T)} = \{ {x}_{\mathbf{A^-}}^{(t+T)}, {y}_{\mathbf{A^-}}^{(t+T)} \} \in \mathbb{R}^{2}$. We represent our probabilistic model as the distribution $p(\mathcal{Y}^t | \mathcal{X}^t)$, where $\mathcal{X}^t = (S_{\mathbf{A^+}}^{(t-H:t)}, \{d_j\}_{j=1}^{N_d})$. For the rest of the paper, we will drop the time $t$ for brevity.



\subsection{Network Architecture}

In this section, we describe our model for predicting the current and future states of an adversary given a history of observations for the blue team agents -- $p(\mathcal{Y} | \mathcal{X})$. Our approach utilizes a combination of mutual information maximization to parameterize the components of a Gaussian mixture model. The architecture consists of four components, 1) encoding the past history of detections ($f_\theta$), 2) encoding the agent positions through a graph neural network ($g_\theta$), 3) mixture of Gaussians decoder ($p_\psi$) and 4) mutual information network ($q_\phi$). Below, we discuss how $f_\theta$ and $g_\theta$ are used to produce an encoded embedding, $e$, and how we enable prediction of adversary location via $p_\psi$ and  $g_\theta$.

\subsubsection{Encoding Detection History}
First, the history of detections $\lbrace d \rbrace_{1..t}$ are encoded in network $f_\theta$. The history of detections does not include agent information and we design this network to focus solely on detections by any agent. Agent positions are passed through network $g_\theta$, which consists of a combination of an LSTM and GNN and \draft{outputs a joint embedding encoding all agent location history}. The LSTM network is used to encode the timeseries information of the blue agent locations, while the GNN is utilized to combine the embeddings from separate agents together. We concatenate the embeddings from the detection network $f_\theta$ and agent position network $g_\theta$ as our encoder embedding $e$.

We compare our models both with and without the agent position network, $g_\theta$, as agent tracks may not always be available to the observers. For example, in the narcotics domain, we may receive detection information from other countries but they may not be willing to share how the information was gathered or where the observer was located.

\subsubsection{Accounting for Multimodality}
\draft{Traditionally, multi-hypothesis tracking is difficult as there is a combinatoric growth for the search space, making it computationally expensive to track all hypotheses \cite{Kim_2015_ICCV}.} To account for this, we use an end-to-end approach learn the parameters of a Gaussian Mixture Model to capture the multimodalities within the tracks. The adversary can be going to a number of locations and the model must account for this uncertainty. Gaussian mixture models are usually parameterized such that the network learns multiple outputs for each mixture component of $\pi_k, \mu_k, \sigma_k^2$ where $k \in G$ and $G$ is the number of Gaussians. 
We instead learn a network, $p_\psi$ that uses a single output for each component, and the mixture component is parameterized by a categorical variable, $\omega$ to produce a bi-variate Gaussian, $\mathcal{Y} \sim \mathcal{N}(\mathbf{\mu}, \mathbf{\Sigma})$ and a weight, $w_k \in W$. We utilize a softmax over the output weights $W$ to produce the mixing coefficients $\pi_k$ for the mixture. \draft{A bi-variate Gaussian is used to capture both the x and y dimension of the opponent's location.} We show empirically that this formulation 1) generalizes better to account for the uncertainty and 2) allows us to utilize a mutual information maximization term to regularize the mixture components. Finally, the posterior mutual information network $q_\phi$ takes as input the bi-variate Gaussian output $\mathcal{Y}, e$ and predicts a categorical distribution over the categorical variable used to parameterize the decoder. We discuss how this network is trained in the next section. 






\subsection{Loss Function}

The total loss function used for training our network is the weighted sum of log-likelihood loss for the tracking of the adversary and the mutual information maximization term. 

\begin{equation} \label{eq:total_loss}
    \mathcal{L} = \mathbb{E} [\text{log} \: p_{\theta}(\mathcal{Y}|\mathcal{X})] - 
    \lambda I(\omega; z, \mathcal{Y}). 
\end{equation}

In information theory, mutual information between $X$ and $Y$, denoted as $I(X; Y)$, measures the amount of shared information learned from one variable with knowledge of the other. This can be defined in terms of the difference in entropy between the two distributions:

\begin{equation}
    I(X;Y) \triangleq H(X) - H(X|Y) = H(Y) - H(Y|X)
\end{equation}

Mutual information maximization has been used in InfoGAN \cite{infoGAN} to produce latent codes that disentangle hidden characteristics within generated images in GANs, and in InfoGAIL \cite{infoGAIL}, where latent \draft{embeddings} were used to capture different styles of demonstrations in an \draft{imitation learning setting}. 

Our novel approach utilizes mutual information maximization to regulate the components of a Gaussian mixture model distribution. We show that this auxiliary objective helps condition the mixture model to better account for the uncertainty and capture the different modalities within trajectories.

We derive an objective function to maximize mutual information by determining a lower-bound below:

\begin{equation}
    \begin{aligned}
        I(\omega; e, \mathcal{Y}) & = H(\omega) - H(\omega| e, \mathcal{Y}) \\
        & = \mathbb{E}_{\omega \sim P(\omega), \mathcal{Y}^t \sim f(z, \omega)} [\text{log} P(\omega | e, \mathcal{Y})]] + H(\omega) \\
        & = \mathbb{E}_{\mathcal{Y} \sim f(z, \omega)} [D_{KL}(P(\omega | e, \mathcal{Y} ) || Q(\hat{\omega} | e, \mathcal{Y}))] + \\
            & \; \; \; \; \; \mathbb{E}_{\omega \sim P(\omega)} [\text{log} (q_\phi(\omega | e, \mathcal{Y})] + H(\omega) \\
        & \geq \mathbb{E}_{\omega \sim P(\omega)} [\text{log} (q_\phi(\omega | e, \mathcal{Y})] + H(\omega)\\
      \end{aligned}
\end{equation}

We parameterize $P(\omega)$ as a uniform categorical distribution where each component ($\omega_k$) is represented as a one-hot categorical variable. The term $\mathbb{E}_{\omega \sim P(\omega)} [\text{log} (q_\phi(\omega | z, \mathcal{Y})$ then reduces to $\sum_{k \in G} p(\omega)  \text{log} (q_\phi(\omega_k | e, \mathcal{Y})$. This can also be interpreted as the cross entropy loss between the predicted $\hat{\omega}_k$ and true $\omega_k$ used to generate $\mathcal{Y}$.

Other researchers have utilized variational inference models such as variational autoencoders and conditional variational autoencoders to learn a latent space of which a generative model can be used to produce multi-modal outputs. However, we find that due to the partially observable nature of our domain, the process of sampling the latent variable and minimizing the log probability of the ground truth location is insufficient for capturing the dynamics of the game. \draft{We therefore require a method for the loss function to explicitly reason about all latent hypotheses rather than sampling.} 


\section{Evaluation}
\label{sec:evaluation}
In this section, we describe our contributed novel opponent modeling domains, the datasets and baselines used to compare our proposed approach, and the metrics used in evaluating the performance of all models.
\subsection{Domains}
We create two challenging multi-agent coordination domains called \textbf{Narco Traffic Interdiction} and \textbf{Prison Escape}, inspired by real-world scenarios, for assessing our proposed approach. These environments are partially observable, and several orders of magnitude larger than previous multi-agent Predator-Prey games \cite{singh2018learning}. We utilize these two diverse domains to show the generalizability of our approach to different opponent modeling problems.


In both domains, a single adversarial opponent is pursued by a team of heterogeneous tracker agents. The goal of the adversary is to reach a hideout \draft{or goal} location while avoiding detection by the tracker agents. The adversary will actively change its \draft{path to the goal or the goal itself} to evade the tracking agents when detected. The goal of the tracker agents is to continuously track and predict the future locations of the adversary. We assume centralized knowledge across agents on the same team, i.e., all tracker agents have knowledge of the other agents and share the adversary detection history. Further, the ability to track the adversary \draft{depends} on the tracking agent type (e.g., an aircraft has a larger detection radius compared to other tracker agents) and is also terrain-dependent. 


\subsubsection{Narco Traffic Interdiction}

This domain simulates illegal maritime drug trafficking on a $7884 \times 3538$ grid along the Central American Pacific Coastline. The adversary, a drug smuggler, is pursued by a team of heterogeneous tracker agents comprising airplanes and marine vessels. Airplanes have a larger search radius and speed than marine vessels, but only the vessels can capture the smuggler. Smugglers must first reach rendezvous points before heading to the hideouts, representing drug handoffs at sea. The locations of hideouts and rendezvous points are unknown to the tracking team. Episodes start after the team learns one location of the smuggler and end when the smuggler reaches a hideout or is captured by law enforcement (similar to other predator-prey environments \cite{singh2018learning}). Future work may incorporate real-time wave data \cite{WaveData}.


\subsubsection{Prison Escape}
In this task, a heterogeneous team of cameras, search parties, and helicopters (blue team) must coordinate to track an escaped prisoner (red team). The game is played on a $2428 \times 2428$ map with varying terrains where each cell on the grid represents the $(x,y)$ location. This domain is motivated by scenarios in military surveillance and border patrol, where there is a need to track and intercept adversarial targets to ensure the safety of the general population. 

\draft{Unlike the Narco domain, where the tracking team has no knowledge of hideout locations, we assume that the blue team has knowledge of a subset of hideouts \emph{a priori} in the Prison Escape scenario. However, the hideouts are sampled at random without replacement from a fixed set for every rollout. We design the Prison Escape scenario to be more challenging than the Narco domain by including more complex evasive behaviors for the prisoner and a fog-of-war that reduces the detection range of the tracking agents. To analyze long horizon predictions, the tracking agents can only \emph{track but not capture the prisoner}. The episode only terminates when either the prisoner has reached a hideout or after maximum timesteps ($T=72 \: \text{hours} = 4320 \: \text{steps}$)}.

\subsubsection{Agent Policies}
We create heuristic policies for both the adversarial agent and also the blue team to collect pairs of blue team observation and adversary states ($\mathcal{X}^t, \mathcal{Y}^t$) to test our prediction algorithms.

\vspace{2pt}
\noindent \textbf{Adversary Policy:}
We create a challenging adversarial policy that attempts to avoid the blue agents and go to unknown hideouts as discretely as possible. The adversary has two main modes of travel 1) traveling to an unknown hideout and 2) active evasion of blue agents. In 1), if the adversary has not detected any of the blue agents near it, it follows an A-Star path planning algorithm to an unknown hideout. The A-Star algorithm minimizes the cost of both the shortest distance to the goal and also the densest forest regions to evade detection from the blue team. In 2), the adversary has detected at least one of the active blue agents. To avoid further detection and tracking, the adversary plans a path to the nearest dark forest to attempt to lose track of the blue team. If unsuccessful for a set amount of time, the adversary plans a path to a known hideout. 
\vspace{2pt}

\noindent \textbf{Blue Team Policy:}
We also create blue team policies that attempt to track and follow the adversary. The blue team has four main modes of tracking 1) converging at a detection point, 2) intercepting along a velocity vector, 3) executing a spiral search pattern, and finally 4) going towards a random location. If there has been a single detection point thus far, all the blue agents will travel from their current location to the detected point. If more detection arrive during this time, the agents will travel along the vector inferred from the last two detections. Finally, if the blue agents did not receive any additional detections, they perform a local search around the last detected location to attempt to find the prisoner. 

\subsection{Datasets} We evaluate \model{} on two \textbf{Narco Traffic Interdiction} datasets - \emph{Smuggler-Low, Smuggler-High} \draft{with opponent detection rates, $13.8\%$ and $31.5\%$}, and three \textbf{Prison Escape} datasets - \emph{Prisoner-Low, Prisoner-Medium, Prisoner-High} \draft{with opponent detection rates, $12.9\%$, $44.0\%$ and $63.1\%$ respectively}. The detection rates are adapted by changing the detection radius of the pursuit agents. For each dataset, we collect 450 rollouts, initialized by different random seeds, and split into train, validation, and test sets as $66.7\%-22.2\%-11.1\%$ respectively. All metrics reported in Tables I and II are evaluated on the held-out test set after training each baseline for 100 epochs with 3 seeds.




\begin{figure}
    \centering
    \includegraphics[width=0.46\textwidth, keepaspectratio]{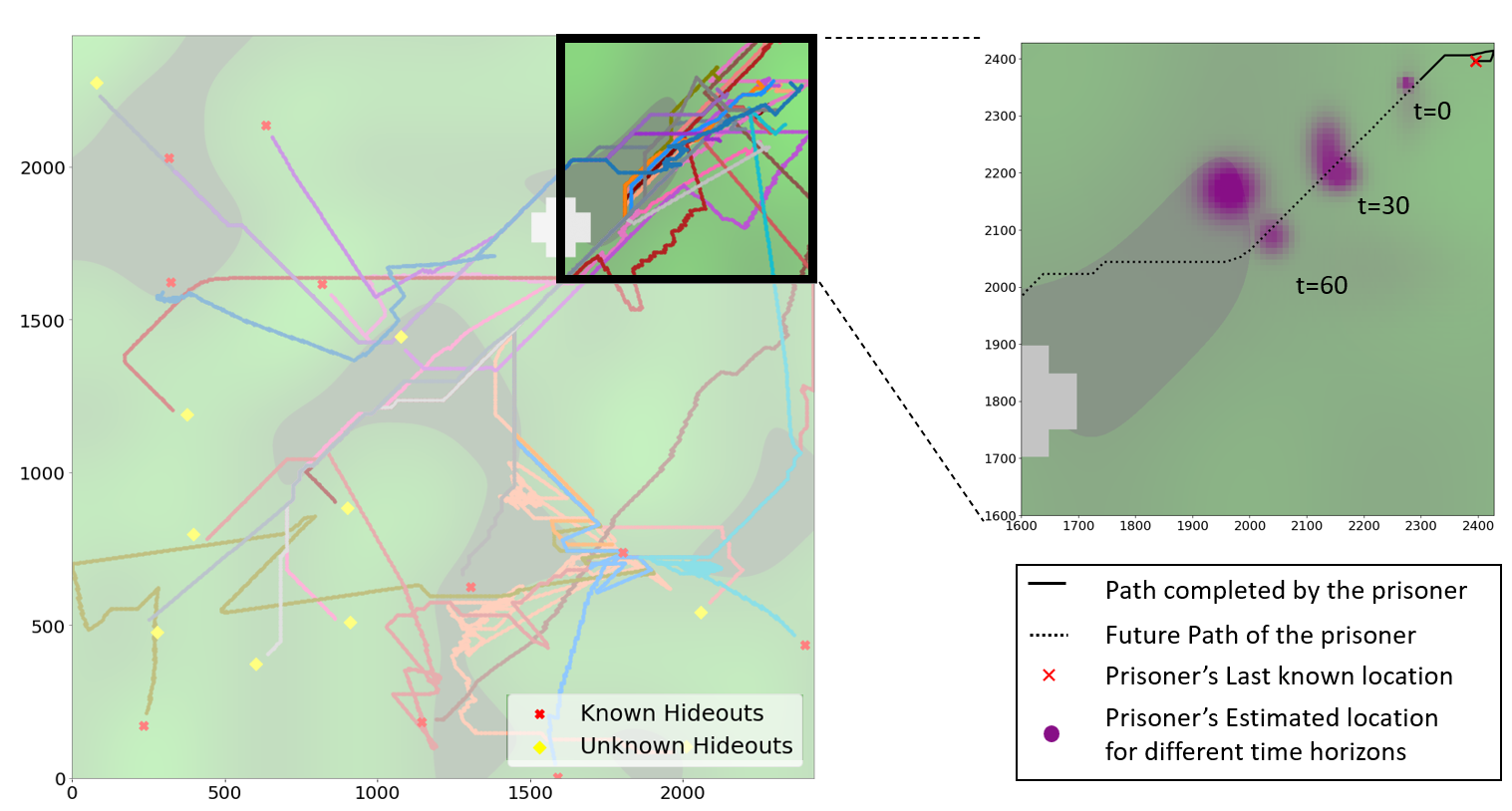}
    
    \caption{Prisoner Escape Scenario: (Left) We show a subset of trajectories taken by the adversarial opponent to go to different hideouts. The darker regions on the map indicate the low visibility range (fog-of-war). (Right) The Mixture of Gaussians 
 for opponent state estimation from \model{} for different prediction horizons}
    \label{fig:paths}
\end{figure}

\subsection{Baselines}
We compare our proposed model against various recurrent neural network configurations with and without a mutual information objective. We implement our method and baselines and report performance after training for 100 epochs \draft{averaged} over 3 random seeds. We note that we do not use other imitation learning baselines such as InfoGAIL as the sparsity of observations makes it difficult to learn a policy for the adversary. 

\begin{enumerate}
    \item LSTM: We test a standard Long-Short Term Memory module where the input is a vector holding all past agent detections. The output is a Gaussian mixture model parameterized such that each component has its own output. 
    
    \item Variational Autoencoder (VAE): We utilize a modified VAE approach inspired by prior work in opponent modeling \cite{papoudakis2020variational}. We modify the approach to only include partial observations of the pursuit team as input (comparable to our other baselines).
    
    \item Variational Recurrent Neural Networks (VRNN): The VRNN was first used for modeling sequential data such as natural speech and handwriting \cite{chung2015recurrent}. Here, VRNN is used as a generative baseline for comparison. The VRNN contains a Variational Autoencoder (VAE) at every timestep to model the dependencies between latent random variables across subsequent timesteps. The VRNN utilizes a conditional prior distribution $z_t$ and  approximate posterior that are both dependent on the prior hidden state $h_{t-1}$.

\end{enumerate}


\subsection{Metrics}
For all the datasets, we assume access to the ground truth (GT) locations of the adversary for evaluation.
\begin{enumerate}
    \item \textbf{Log-Likelihood (LL):} The mean LL of the GT adversary location under the  multi-modal output distribution generated by the model.
    \item \textbf{Average Displacement Error (ADE):} The mean $l_2$ distance between the GT and the predicted adversary location.
    \item \textbf{Confidence Threshold (CT$_\delta$):} The fraction of time steps for which the model predicts that the opponent is within a distance $\delta$ from the GT location with probability greater than some threshold (say $p \geq 0.5$).
\end{enumerate}

\section{Results and Discussions}
\label{sec:results}






We evaluate the performance of \model{} on the two domains: Narco Traffic Interdiction and Prisoner Escape. We benchmark our approach against several state-estimation baselines including 1) VRNNs \cite{chung2015recurrent} and 2) Opponent VAEs \cite{papoudakis2020variational}. We also provide ablations for our model with and without 1) mutual information maximization, 2) agent position GNN encoder, and 3) structure of a categorical variable for the Gaussian mixture model. We train 3 models with random seeds for every condition and time horizon and average the results. We show our results in Table \ref{tab:prisoner} and \ref{tab:smuggler} for the two domains, spanning three different time horizons and five different detection rates. We find that our models (ablations with and without MI and GNN) outperform all baselines on log likelihood, ADE, and confidence threshold. We denote the way we paramaterize the mixture of Gaussians as $\omega$MM for $\omega$ Mixture Model.

\input{prisoner_table}

\subsection{Mutual Information Impact}


We find that mutual information posterior increases the log-likelihood and confidence threshold while lowering the ADE, especially when predicting timesteps close to the present. The mutual information term has a smaller effect when predicting further into the future, likely because of the problem's increased difficulty. Our models outperform Variational models because they explicitly account for mixture components in the negative log-likelihood loss, limiting the number of possible outputs the decoder can use and reasoning over the multi-modal hypothesis space better. Variational methods have worked well in high-density, fully observable filtering and prediction domains, but we find in our domains that these sparse detections do not provide enough information to create a useful posterior distribution. 



\subsection{Ablation Study}
\subsubsection{Utilizing a Categorical Variable for the Mixture of Gaussians}
In this ablation, we compare the LSTM model to our model without mutual information and the GNN (Ours w/o GNN \& MI). These two models have the same agent history detection encoder but different parameterizations for the Gaussian mixture model. The LSTM model utilizes multiple outputs for each mixture component, while our model uses a single output for multiple mixture components. We find that even with the same loss function, our model drastically outperforms the LSTM model. We theorize that this formulation may lead to better performance due to the shared parameters used within the decoder. 

\subsubsection{Agent Position Encoder}
We analyze how the GNN encoder affects the model (Ours vs Ours w/o GNN). We compare results with and without agent tracks as agent tracks may not always be available to the observers. We hypothesized that encoding a history of searching agent locations can help the models generalize, as this information informs the models where the opponent has \textbf{not} been in previous timesteps. We find that including the agent tracks produces better results for the Smuggler-High domain  on predicting all timestep horizons, but results are mixed regarding the other datasets where the agent encoder improves the log-likelihood on some of the higher detection-rate datasets (Prisoner-Mod, Prisoner-High). Because our domains have such a large search space, the amount of information gained from paths where no detection has occurred is small. In smaller domains, the information gained by occupying a state and viewing surrounding states is a much larger percentage. This may be the reason why the datasets with higher detection radii in Prisoner-Mod and Prisoner-High achieve better performance with the agent encoder.

\input{smuggler_table}

\section{Limitations and Future Work}
Our work has several limitations: We are currently unable to generate predictions for multiple time horizons. Additionally, more advanced agent heuristics could be used to represent a distribution of strategies to make the filtering and prediction tasks harder and further evaluate the efficacy of our model. Reinforcement learning could be used to provide new evasive behaviors covering a wider range of strategies. Future work includes encoding the terrain information regarding detection ranges to better capture the evasive behaviors of the adversary. We can additionally account for the error uncertainty in sensors with sensor fusion for the adversary's state estimation. 

\section{Conclusion}

In conclusion, our proposed approach, \model{}, for adversarial opponent tracking in large state spaces, utilize a deep learning architecture that predicts the current and future states of an adversarial agent from partial observations of its trajectory. Our approach shows significant improvement over variational methods and demonstrates its generalizability through two contributed open-source domains, Narco Traffic Interdiction and Prison Escape.

\bibliographystyle{plain}

\end{document}

%% file: prisoner_table.tex
\setlength{\tabcolsep}{3pt}
\newcommand{\ra}[1]{\renewcommand{\arraystretch}{5}}
\begin{table*} \centering

\vspace{0.2cm}
\begin{tabular}{@{}lrrr|rrr|rrr|rrr@{}}
\toprule
Prisoner-Low \\
\toprule
& \multicolumn{3}{c}{} & \multicolumn{3}{c}{Log-Likelihood}& \multicolumn{3}{c}{$ADE$}& \multicolumn{3}{c}{CT$_\delta$} \\
\cmidrule{5-7} \cmidrule{8-10} \cmidrule{11-13}
& GNN  & $\omega$MM & MI & $0$ min & $30$ min & $60$ min & $0$ min & $30$ min & $60$ min & $0$ min & $30$ min & $60$ min \\ \midrule

LSTM                & -          & -          & -          & 4.901          & 4.135          & 2.843          & 0.069          & 0.090          & 0.136          & 0.910          & 0.844          & 0.728          \\
VRNN Seq            & -          & -          & -          & 4.421          & 4.217          & 3.850          & 0.106          & 0.093          & 0.119          & 0.738          & 0.677          & 0.610          \\
VAE Opponent        & -          & -          & -          & 5.121          & 3.973          & 3.685          & 0.085          & 0.095          & 0.119          & 0.723          & 0.600          & 0.561          \\
Ours (w/o GNN \& MI) & -          & \checkmark & -          & 5.784          & 5.504          & 4.460          & \textbf{0.060} & 0.083          & \textbf{0.109} & 0.958          & 0.916          & 0.891          \\
Ours (w/o GNN)      & -          & \checkmark & \checkmark & \textbf{6.381} & \textbf{5.606} & \textbf{4.702} & \textbf{0.060} & \textbf{0.080} & 0.110          & \textbf{0.960} & \textbf{0.925} & \textbf{0.902} \\
\midrule
\textbf{Includes Agent Position} \\
Ours (w/o MI)       & \checkmark & \checkmark & -          & 4.609          & \uline{5.071}    & 4.117          & 0.062          & 0.082          & 0.112          & \uline{0.960}    & \uline{0.926}    & \uline{0.876}    \\
Ours                & \checkmark & \checkmark & \checkmark & \uline{5.230}    & 4.845          & \uline{4.789}    & \uline{0.061}    & \uline{0.081}    & \uline{0.107}    & 0.957          & 0.931          & 0.857         \\

\bottomrule
\bottomrule

\toprule
Prisoner-Medium \\
\toprule
& \multicolumn{3}{c}{} & \multicolumn{3}{c}{Log-Likelihood}& \multicolumn{3}{c}{$ADE$}& \multicolumn{3}{c}{CT$_\delta$} \\
\cmidrule{5-7} \cmidrule{8-10} \cmidrule{11-13}
& GNN  & $\omega$MM & MI & $0$ min & $30$ min & $60$ min & $0$ min & $30$ min & $60$ min & $0$ min & $30$ min & $60$ min \\ \midrule

LSTM                & -          & -          & -          & 5.858                     & 4.091          & 2.836          & 0.066                     & 0.088          & 0.125          & 0.918                     & 0.830          & 0.618          \\
VRNN Seq            & -          & -          & -          & \multicolumn{1}{r}{3.732} & 4.214          & 3.599          & \multicolumn{1}{r}{0.172} & 0.086          & \textbf{0.110} & \multicolumn{1}{r}{0.495} & 0.666          & 0.553          \\
VAE Opponent        & -          & -          & -          & \multicolumn{1}{r}{5.432} & 4.372          & 3.281          & \multicolumn{1}{r}{0.063} & 0.080          & 0.117          & \multicolumn{1}{r}{0.734} & 0.665          & 0.445          \\
Ours (w/o GNN \& MI) & -          & \checkmark & -          & 7.214          & 5.065          & 4.256          & \textbf{0.047} & 0.078          & \textbf{0.110} & \textbf{0.973} & 0.939          & 0.883          \\
Ours (w/o GNN)       & -          & \checkmark & \checkmark & \textbf{7.981} & \textbf{5.288} & \textbf{4.385} & 0.049          & \textbf{0.077} & \textbf{0.110} & 0.965          & \textbf{0.952} & \textbf{0.901} \\
\midrule
\textbf{Includes Agent Position}\\
Ours (w/o MI)        & \checkmark & \checkmark & -          & 7.026          & 4.970          & \uline{4.109}    & \uline{0.048}    & 0.078          & \uline{0.113}    & 0.969          & 0.922          & 0.882          \\
Ours                 & \checkmark & \checkmark & \checkmark & \uline{8.406}    & \uline{5.270}    & 4.059          & 0.049          & \uline{0.073}    & \uline{0.113}    & \uline{0.972}    & \uline{0.943}    & \uline{0.898}  \\

\bottomrule
\bottomrule

\toprule
Prisoner-High \\
\toprule
& \multicolumn{3}{c}{} & \multicolumn{3}{c}{Log-Likelihood}& \multicolumn{3}{c}{$ADE$}& \multicolumn{3}{c}{CT$_\delta$} \\
\cmidrule{5-7} \cmidrule{8-10} \cmidrule{11-13}
& GNN  & $\omega$MM & MI & $0$ min & $30$ min & $60$ min & $0$ min & $30$ min & $60$ min & $0$ min & $30$ min & $60$ min \\ \midrule

LSTM                & -          & -          & -          & 6.955                     & 4.110          & 2.555                     & 0.042                     & 0.059          & 0.111                     & 0.946                     & 0.823          & 0.528                     \\
VRNN Seq            & -          & -          & -          & \multicolumn{1}{r}{5.037} & 4.345          & 2.967                     & \multicolumn{1}{r}{0.105} & 0.059          & 0.100            & \multicolumn{1}{r}{0.714} & 0.731          & 0.538                     \\
VAE Opponent        & -          & -          & -          & 2.991 & 4.172          & 2.754 & 0.200 & \textbf{0.054}          & 0.123 & 0.528 & 0.657          & 0.382 \\

Ours (w/o GNN \& MI) & -          & \checkmark & -          & 8.515  & 4.422          & \textbf{3.297} & 0.016 & 0.057 & 0.095 & \textbf{0.965} & 0.923          & \textbf{0.803} \\
Ours (w/o GNN)      & -          & \checkmark & \checkmark & \textbf{10.862} & \textbf{4.535} & 3.221          & \textbf{0.015} & 0.056          & \textbf{0.092} & 0.953 & \textbf{0.932} & 0.801          \\
\midrule
\textbf{Includes Agent Position}\\

Ours (w/o MI)       & \checkmark & \checkmark & -          & 8.346           & \uline{4.518}    & 3.147          & 0.018          & 0.049          & 0.098          & 0.965          & 0.892          & 0.753          \\
Ours                & \checkmark & \checkmark & \checkmark & \uline{11.094}    & 4.389          & 3.202          & \uline{0.015}    & \uline{0.046}    & \uline{0.092}    & \uline{0.970}    & \uline{0.910}          & \uline{0.795}           \\

\bottomrule
\end{tabular}

\caption{Prison Escape Environment Results for different prediction horizons. Underlined and bolded indicate the best models with and without agent positions, respectively}

\label{tab:prisoner}
\end{table*}

%% file: smuggler_table.tex
\begin{table*} \centering
\vspace{0.2cm}
\begin{tabular}{@{}lrrr|rrr|rrr|rrr@{}}
\toprule
Smuggler-Low \\
\toprule
& \multicolumn{3}{c}{} & \multicolumn{3}{c}{Log-Likelihood}& \multicolumn{3}{c}{$ADE$}& \multicolumn{3}{c}{CT$_\delta$} \\
\cmidrule{5-7} \cmidrule{8-10} \cmidrule{11-13}
& GNN  & $\omega$MM & MI & $0$ min & $30$ min & $60$ min & $0$ min & $30$ min & $60$ min & $0$ min & $30$ min & $60$ min \\ \midrule

LSTM                & -          & -          & -          & 3.241          & 3.453          & 3.036          & 0.166          & 0.155          & 0.163          & 0.788          & 0.795          & 0.741         \\
VRNN Seq            & -          & -          & -          & 4.455          & 3.704          & 3.303          & 0.147          & 0.156          & 0.186 & 0.617          & 0.611          & 0.473         \\
VAE Opponent        & -          & -          & -          & 3.962          & 3.410          & 3.129          & 0.137          & 0.151          & 0.187          & 0.671          & 0.541          & 0.560         \\


Ours (w/o GNN \& MI) & - & \textbf{\checkmark} & - & \textbf{4.941} & \textbf{3.562} & \textbf{3.601} & 0.122 & \textbf{0.142} & \textbf{0.159} & \textbf{0.874} & \textbf{0.876} & \textbf{0.89} \\
Ours (w/o GNN)      & -                   & \checkmark          & \checkmark          & 4.817 & 3.208          & 3.557          & \textbf{0.121} & 0.144          & 0.181          & 0.848          & 0.862          & 0.86          \\
\midrule
Ours (w/o MI)       & \checkmark    & \checkmark    & -                   & \uline{4.809}    & \uline{3.808}    & 3.41           & \uline{0.127}    & 0.148    & 0.159          & 0.892          & \uline{0.86}     & 0.854         \\
Ours                & \checkmark                   & \checkmark          & \checkmark             & 4.725    & 3.529    & \uline{3.645}    & \uline{0.127}    & \uline{0.145}    & \uline{0.162}    & \uline{0.895}    & 0.855          & \uline{0.866}  \\

\bottomrule
\bottomrule

\toprule
Smuggler-High \\
\toprule

LSTM                & -          & -          & -          & 2.927          & 3.252          & 2.709          & 0.162          & 0.152          & 0.167          & 0.746          & 0.77           & 0.733          \\
VRNN Seq            & -          & -          & -          & 3.770          & 3.157          & 3.361          & 0.138          & 0.153          & 0.183 & 0.673          & 0.561          & 0.560          \\
VAE Opponent        & -          & -          & -          & 3.656          & 3.088          & 2.937          & 0.138          & 0.146          & 0.168          & 0.612          & 0.539          & 0.564          \\

Ours (w/o GNN \& MI)       & -          & \checkmark & - & 4.63           & 3.895          & 3.586          & \textbf{0.125} & \textbf{0.144}         & \textbf{0.161}       & \textbf{0.829} & \textbf{0.861} & \textbf{0.856} \\
Ours (w/o GNN) & -          & \checkmark & \checkmark         & \textbf{5.194} & \textbf{4.282} & \textbf{3.681} & 0.131          & 0.163         & 0.174       & 0.825          & 0.843          & 0.83           \\
\midrule
\textbf{Includes Agent Position} \\
Ours (w/o MI)                & \checkmark & \checkmark & - & 4.279          & 3.628          & \uline{3.437}          & 0.126          & 0.143         & 0.158       & 0.825          & 0.815          & \uline {0.786} \\
Ours        & \checkmark & \checkmark & \checkmark          & \uline{4.678}    & \uline{3.814}          & 3.133          & \uline{0.123}    & \uline{0.14} & \uline{0.156} & \uline{0.836}    & \uline{0.822}    & 0.773          \\

\bottomrule
\end{tabular}
\caption{Narco Traffic Interdiction Environment Results for different prediction horizons. Underlined and bolded indicate the best models with and without agent positions, respectively}

\label{tab:smuggler}
\end{table*}